\documentclass[preprint,12pt]{elsarticle}
\usepackage{amsmath,amssymb,amsthm}
\usepackage{hyperref}

\newtheorem{theorem}{Theorem}
\newtheorem{proposition}{Proposition}
\journal{Information Processing Letters}

\begin{document}

\begin{frontmatter}

\title{Dual Computational Horizons: Incompleteness and Unpredictability in Intelligent Systems}

\author{Abhisek Ganguly\corref{cor1}}
\cortext[cor1]{Corresponding author}
\ead{abhisek@jncasr.ac.in}
\address{Engineering Mechanics Unit, Jawaharlal Nehru Centre for Advanced Scientific Research, Bangalore 560064, India}

\begin{abstract}
We formalize two independent computational limitations that constrain algorithmic intelligence: formal incompleteness and dynamical unpredictability. The former limits the deductive power of consistent reasoning systems while the latter bounds long-term prediction under finite precision. We show that these two extrema together impose structural bounds on an agent’s ability to reason about its own predictive capabilities. In particular, an algorithmic agent cannot verify its own maximal prediction horizon universally. This perspective clarifies inherent trade-offs between reasoning, prediction, and self-analysis in intelligent systems. The construction presented here constitutes one representative instance of a broader logical class of such limitations. 
\end{abstract}

\begin{keyword}
Algorithmic intelligence \sep Computability \sep Gödel incompleteness \sep Lyapunov exponent \sep Prediction horizon \sep Computational limits
\end{keyword}

\end{frontmatter}

%-------------------------------------------------
\section{Introduction}

Modern AI has seen remarkable applications with groundbreaking results, yet fundamental theoretical limits remain underexplored. While scaling laws and architectures dominate research, intrinsic computational constraints from logic and dynamical systems place strict bounds on algorithmic intelligence. We identify and link two independent horizons:
\begin{enumerate}
    \item \textbf{Formal incompleteness:} No consistent system capable of arithmetic can be deductively complete \cite{godel,turing}.
    \item \textbf{Dynamical unpredictability:} Deterministic systems with positive Lyapunov exponents admit finite prediction horizons under finite-precision observations \cite{ott,cohen2021}.
\end{enumerate}
We formalize their joint implications for algorithmic agents and connect them to phenomena such as multi-step reasoning limitations.

\paragraph{Related Work} Gödel incompleteness puts constraints on the formal reasoning in AI \cite{godel,turing,bereska2024}, while chaos theory establishes limits on long-term prediction \cite{ott,cohen2021}. Computational irreducibility further restricts algorithmic shortcuts \cite{wolfram}. Prior work often considers these separately; we explicitly link them in a unified computational framework. We do not claim new incompleteness results or new bounds in dynamical systems. We rather formalize their joint implications for algorithmic agents.

%-------------------------------------------------
\section{Formal horizon: Limits of deductive reasoning}
Let us take an example reasoning system, which can be represented as a pair $(\mathcal{S}, \vdash)$ where $\mathcal{S}$ is a recursively enumerable theory over a countable language $\mathcal{L}$ and $\vdash$ denotes effective derivability. We assume $\mathcal{S}$ is consistent and capable of representing arithmetic.

\begin{theorem}[Gödel's First Incompleteness Theorem~\cite{godel}]
Let $\mathcal{S}$ be consistent and recursively enumerable. Then $\mathcal{S}$ is incomplete if there exists $\varphi \in \mathcal{L}$ such that neither $\mathcal{S} \vdash \varphi$ nor $\mathcal{S} \vdash \neg \varphi$.
\end{theorem}

This suggests that any agent whose reasoning is formalizable as a computable procedure cannot achieve deductive closure. In particular, such an agent cannot internally verify all truths expressible in its own language. This inherent limit is independent of the compute power of the agent.

%-------------------------------------------------
\section{Dynamical horizon: Finite prediction}

Let $(X, d)$ be a metric space in $n$ dimensions, $X \subset \mathbb{R}^n$, and let
\begin{equation}
x_{t+1} = F(x_t),
\end{equation}
be a deterministic dynamical system with maximal Lyapunov exponent $\lambda>0$. Suppose initial conditions are known only up to finite precision $\epsilon>0$, and let there be a computable predictor with an error $\epsilon$, given by $P_{\epsilon}$ such that:
\begin{equation}
P_\epsilon:X \to X.
\end{equation}
Note that the predictor $P$ is a finite computational representation of the dynamical system $F$.
Let us consider $d$ as a distance metric in the phase space for the predicted and true states given by $P_\epsilon^t(x)$ and $F^t(x)$ respectively, at time $t$. Then, the prediction error grows at least as
\begin{equation}
d( P_\epsilon^t(x),F^t(x)) \ge C \epsilon e^{\lambda t}.
\end{equation}
For a fixed tolerance $\delta>0$, this lets us define the \emph{prediction horizon}:
\begin{equation}
T(\epsilon) := \frac{1}{\lambda}\log\frac{\delta}{C\epsilon}.
\end{equation}
Therefore, no algorithm can maintain arbitrary long-term prediction beyond $T(\epsilon)$. While this is a well understood computational limitation, the derived consequences start from here.
%-------------------------------------------------
\section{Uncomputable self-prediction}

Let us model an algorithmic agent~\cite{google2025,aws2025,ruffini2024} as $A=(R,P,M)$ with reasoning $R$, predictor $P$, and internal model $M$. Here, $R \subseteq \mathcal{S}$, where $\mathcal{S}$. In the agent $A$, \textit{partial} representations of $R$ and $P$ are stored in $M$ using a countable language $\mathcal{L}$. We show why it is \textit{partial} as a corollary to the dual limitations. 

Let
\begin{equation}\label{eq:maximal_time}
T_A(\epsilon) := \text{maximal time $A$ can predict accurately.}
\end{equation}
\begin{proposition}[Dual-Horizon Limitation]
Let $R$ be a consistent, recursively enumerable reasoning system. Then $R$ cannot compute $T_A(\epsilon)$ for arbitrary deterministic environments with $\lambda>0$.
\end{proposition}

\subsection{Proof sketch:} 
To compute $T_A(\epsilon)$, Eq.~\ref{eq:maximal_time} can be represented as:
\begin{equation}\label{eq:tmax_comp}
    T_A(\epsilon) = \sup\{ t \mid \forall x \in X, \quad \forall \tau \leq t, \quad d(F^\tau(x), P_\epsilon^\tau(x)) < \delta \}
\end{equation}

To compute Eq.~\ref{eq:tmax_comp}, the reasoner $R$ must decide, for every $t$, whether its predictor $P$ (parameterized by internal model $M$) satisfies the error bound over all initial states. 

\subsubsection{The halting problem}
For hyperbolic systems ($\lambda > 0$), this requires simulating $P^t(x)$ to arbitrary limits. Formally, the set $\{ t \mid R \vdash (P^t \approx F^t) \}$ is not computable, as its complement relates to the halting problem: $R$ cannot decide if divergent simulations ``halt within $\delta$'' \cite{turing}. Thus $R \not\vdash T_A(\epsilon)$ for arbitrary generality.

\subsubsection{The shadowing lemma}
While not a primary limiting pillar, this aspect needs some discussion. The exponential error growth $d(F^t(x), P_\epsilon^t(x)) \geq C\epsilon e^{\lambda t}$ implies no finite computation is sufficient in order to claim the existence/non-existence of $\sup\{t\}$ accurately by the agent, for $t > T(\epsilon)$ owing to the shadowing lemma~\cite{ott,hammel1987}.

Suppose the agent finds a $\sup\{t\}$ for a numerical (here, numerical means using the reasoning $R$, in language $\mathcal{L}$) trajectory originating from $\hat{x}_i$, i.e., the finite encoding of $x_i$ inside the agent. It would mean that there is a real shadowing trajectory from a different real initial condition $x'_i$ that follows this numerical trajectory. While this establishes the credibility of the supposedly calculated trajectory, albeit from a different real starting condition, it leads to another problem for the agent. It is unprovable that this shadowing trajectory will not blow up \textit{past} $\sup\{t\}$ with finite computation. Hence, the claim by the agent would be wrong in terms of accuracy or provability.

\subsubsection{Gödel's second incompleteness}
Moreover, $M$ is the system that encodes $R$'s own dynamics, albeit not completely. $M$ cannot contain \textit{all} the statements about $R$, i.e., all the performance proofs about the encoded reasoning system. We may write:
\begin{gather}
    r, p \subseteq M,  \notag \\
    r \subset R, p \subset P.
\end{gather}
Therefore, the verification of a particular statement given as,
\begin{equation}
    \forall \,x \in X, \forall \, \tau \in t, \, P^\tau_\epsilon(x) \approx F^\tau(x), 
\end{equation}
 leads to self-reference. Deciding this statement is a $\Pi_1^0$ statement~\cite{enderton} in arithmetic. It is then equivalent to saying: ``$\forall$ (For all) trajectories, error remains bounded''. By Gödel's second incompleteness theorem, consistent $R$ cannot prove its own consistency, hence is not guaranteed to uniformly verify such simulation statements across all $t$.  
 
 For an agent $A = (R,P,M)$, there exist an infinite number of combinations of $R,P,M$ owing to the expected complexity of advanced algorithmic intelligence. For atleast one of such combination is bound to have this statement as a true, unprovable $\Pi_1^0$ statement.\\

% \noindent\textbf{Remark:}
% The above argument is intended as a structural limitation rather than a complete computability-theoretic proof. 
% It is important to note here that neither incompleteness nor unpredictability due to chaos alone points towards Proposition 1. The limitation arises when an agent attempts to formally verify the bounds on its own predictive competence.\\

\subsection{Illustrative Example: Self-Prediction of a Chaotic Simulator}
Consider an agent $A$ whose environment model $M$, encoding the deterministic chaotic system:
\begin{equation}
x_{t+1} = F(x_t),
\end{equation}
with maximal Lyapunov exponent $\lambda>0$, implemented internally as part of its world simulator. Suppose $A$ attempts to compute its own prediction horizon $T_A(\epsilon)$ for this problem by simulating future states of $M$ using finite-precision arithmetic. It gives out an answer $t_{max}$. But since this is a dynamical system that is being encoded, shadowing lemma tells us that $t_{max}$ is not verifiable by the agent as the right answer. Furthermore, if the problem doesn't have any $t_{max}$, the agent will not be able to determine it as well as it is not determinable as it becomes essentially a halting problem. The agent eventually stops which is being governed by the timeout systems in $R$ as a part of it's reasoning of human command to not use infinite resources over time. This is a Turing limitation.
This problem reduces to undecidable self-verification when $M$ includes $R$ itself, as it essentially becomes a task of reasoning for $R$, \textit{about} itself and using the tools \textit{only} $R$ ha. Note that the model $M$ has a limited true information about $R$ as well. Thus, algorithmic computation of $T_A(\epsilon)$ by $A$ is not possible universally.

Furthermore, the present encoded networks themselves have internal numerical scheme instabilities. This is not surprising as the internal flow of information can be modeled neatly by continuous-depth neural networks and Neural Ordinary Differential Equations. These equations being non-linear, are therefore not alien to stability analysis in the light of dynamical systems either~\cite{haber2017stable}.

The agent does not merely fail to know the true prediction horizon; it also fails to know whether it has the access to such knowledge in principle.

\section{Discussion}

The dual-horizon perspective highlights a trade-off between stability of predictions and expressive/complex computational behavior. Systems operating near regimes of high sensitivity to initial conditions may achieve greater representational or optimization capacity, but at the cost of reduced long-term predictability. Conversely, constraining dynamics to improve predictability may limit expressivity.

From this viewpoint, phenomena such as error accumulation in long chains of reasoning may be understood as manifestations of finite prediction limits rather than implementation flaws. Modern neural language models, which operate as large-scale predictors under finite precision, provide an illustrative example: beyond certain depths, uncertainty accumulates and outputs may drift from intended linguistic outputs. We emphasize that this interpretation is illustrative rather than empirical.

Finally, incompleteness points towards inherent limits on self-verification. For sufficiently expressive reasoning systems, certain internal states cannot be conclusively classified using the system’s own rules. This aligns with recent analyses suggesting that absolute formal guarantees of safety or correctness may be unattainable for advanced autonomous agents, motivating probabilistic or external validation approaches \cite{vassilev2025}. 

This discussion essentially points to the identification of Turing type limitation which belongs to the class of $\Pi_1^0$ type undecidability. Other forms of undecidability may exist in different logical classes as well and may be explored. 

\section{Conclusion}

We identified two fundamental computational horizons, namely formal incompleteness and finite prediction under dynamical instability, and showed that their interaction imposes intrinsic limits on algorithmic intelligence. We present an argument that in general, no agent with computable reasoning and finite-precision prediction can determine its own maximal prediction limit. This result may act as a step towards the clarification on why long-horizon reasoning and self-verification inevitably degrade, even for highly capable systems. These limitations are structural rather than architectural and persist regardless of scale or implementation. We show that algorithmic intelligence does not overcome the bounds placed by incompleteness and chaos, but rather functions within these bounds. This approach possibly provides us a principled, foundational basis for analyzing predictability, interpretability, and safety of modern and post-modern algorithmically intelligent systems.

%-------------------------------------------------
\bibliographystyle{elsarticle-num}
\bibliography{biblio}

\end{document}